\documentclass[letterpaper, 10 pt, conference]{ieeeconf}
\IEEEoverridecommandlockouts
\usepackage{graphicx}
\usepackage{cite}  
\usepackage{amsmath,amssymb}
\usepackage{bm}
\usepackage{dsfont}
\usepackage{tabularx}
\usepackage{array}
\usepackage{makecell}
\usepackage{booktabs}
\usepackage{multirow}
\usepackage{comment}

\usepackage{algorithm}
\usepackage{algorithmicx}
\usepackage{algpseudocode}
\usepackage{wrapfig}
\usepackage{subcaption}

\usepackage{siunitx}
\usepackage{threeparttable}
\usepackage{adjustbox}
\usepackage{xcolor}
\usepackage{pifont}

\newcolumntype{N}{>{\centering\arraybackslash}m{1.15cm}}
\newcolumntype{L}[1]{>{\raggedright\arraybackslash}p{#1}}
\newcolumntype{C}[1]{>{\centering\arraybackslash}p{#1}}
\newcolumntype{Y}{>{\raggedright\arraybackslash}X}

\algrenewcommand\algorithmicindent{1.2em}

\usepackage[table]{xcolor}
\usepackage{hyperref}

\title{Robot Motion Transition with a Human Motion Prior}

\author{Taehyun Byun$^{1}$, Taemoon Jeong$^{1}$, Yoonbyung Chai$^{1}$, Woojin Wie$^{2}$, Insu Park$^{2}$, Kyungjae Lee$^{3}$, and Sungjoon Choi$^{1}\textsuperscript{*}$%
\thanks{${^1}$Taehyun Byun, Taemoon Jeong, Yoonbyung Chai and Sungjoon Choi are with the Department of Artificial Intelligence, Korea University, Seoul, Republic of Korea. (e-mails: \{taehyun-byun, taemoon-jeong, yoonbyung-chai, sungjoon-choi\}@korea.ac.kr)}
\thanks{${^2}$ Woojin Wie and Insu Park is with ROBOTIS, (e-mail: \{wwj,pis\}@robotis.com)}
\thanks{${^3}$ Kyungjae Lee is with the Department of Statistics, Korea University, Seoul, Republic of Korea. (e-mail: kyungjae\_lee@korea.ac.kr)}
}

\hypersetup{hidelinks}

\begin{document}
\maketitle
\thispagestyle{empty}
\pagestyle{empty}

\begin{abstract}
Humanoid motion-tracking policies can reproduce diverse behaviors, yet connecting independently acquired motions into a continuous sequence remains challenging. Interpolation connects boundary poses without considering whether their relative position, heading, and timing provide suitable conditions for a transition. A generative motion prior can produce richer motion, but poorly arranged boundaries may cause it to generate excessive locomotion, turning, or another unintended action. We present Spatio-Temporal Alignment with a Motion Prior (STAMP), a framework for connecting independently acquired humanoid motions through a Robot-to-Human-to-Robot process. STAMP uses human motion space to draw on rich motion priors and reuse transition knowledge across robot embodiments. Its key component is a learned alignment model that determines the boundary conditions for motion inbetweening. Given the two motions retargeted to a human rig, the alignment model ranks candidate positions, headings, and transition durations for the following motion. A human motion prior generates the transition under the selected alignment.
The completed sequence is retargeted to the robot and used as a reference to train a motion-tracking policy. We evaluate the tracking success of the generated transitions on Unitree~G1 and~ROBOTIS K1. Physical robot demonstrations further show that STAMP connects independently acquired motions into long-horizon sequences. Project page: \url{https://childtoy.github.io/STAMP-project}
\end{abstract}

\section{Introduction}\label{sec:introduction}
Recent advances in physics-based motion tracking enable humanoid robots to reproduce diverse and dynamic motions~\cite{peng2018deepmimic,peng2021amp,beyondmimic}. However, tracking individual motions and connecting independently acquired motions into a continuous sequence are different problems.

Robot motions are commonly generated and validated as separate clips. When two clips are connected, the position and heading of the following motion and the time available for the transition are not given. Two motions can each be plausible on their own but difficult to connect under an arbitrary relative placement. These spatial and temporal choices determine the body movement required between their boundaries.

A simple approach is to return the robot to a predefined pose between motions, but this stop-and-restart behavior interrupts the motion flow. Another approach is to interpolate directly between the two boundary poses. However, directly interpolating boundary poses alone does
not account for the incoming and outgoing motion or foot-contact changes. For a humanoid, a change in the support foot may require a step or a transfer of body weight that pose interpolation does not explicitly model.

Recent approaches handle humanoid motion transitions through graph-based motion switching, motion matching, or policy blending~\cite{lau2026switch,wu2026perceptive,rpg_2026}. These methods transition at intermediate states selected for compatibility, or blend policies around the switching point. As a result, the constituent motions may be truncated or altered, and the content of each motion is not necessarily preserved. STAMP instead connects the end of the preceding motion to the start of the following motion while preserving both input clips.

Motion inbetweening generates missing motion between given boundary
conditions~\cite{harvey2020robust,kim2022conditional,cohan2024flexible}.
Human motion data provide rich examples of body movements and contact
changes for learning such transitions~\cite{bonesseed,kimodo_2026}.
MotionBricks further demonstrates the use of motion inbetweening for
real-world humanoid control~\cite{motionbricks_2026}.

However, inbetweening boundaries extracted from a continuous motion
already have a defined spatial and temporal relationship. Independently
acquired motions do not provide this relationship. Applying a motion
prior to an arbitrary arrangement may require excessive locomotion,
turning, or another unintended action to connect the motions. Connecting such motions therefore requires selecting suitable boundary conditions before generating the transition.

To address this problem, we propose Spatio-Temporal Alignment with a Motion Prior (STAMP), a Robot-to-Human-to-Robot framework that connects robot motions through a shared human motion space. STAMP learns from continuous human motion to rank candidate positions, headings, and transition durations, then uses a human motion prior to generate a transition under the selected conditions. The completed sequence is retargeted to the target robot and used to train a robot-specific motion-tracking policy. Experiments on Unitree~G1 and ROBOTIS~K1 evaluate the trackability of the resulting transitions, and physical robot demonstrations show their use in longer motion compositions.

Our contributions are summarized as follows.

\begin{itemize}
    \item We introduce STAMP, a Robot-to-Human-to-Robot framework
    that uses human motion knowledge to connect independently
    acquired humanoid motions across different robot embodiments.

    \item We learn a spatial and temporal alignment model from
    continuous human motions and counterfactual boundary arrangements.
    The model ranks candidate positions, headings, and transition
    durations to select boundary conditions for motion inbetweening.

    \item We evaluate robot tracking success on Unitree~G1 and ROBOTIS~K1 against
    interpolation and generative motion baselines, analyze alignment
    selection through a ranking ablation, and demonstrate long-horizon
    motion compositions on the ROBOTIS~K1.
\end{itemize}

\section{Related Work}
\label{sec:related}

\subsection{Humanoid Motion Tracking and Transition}

Physics based imitation learning enables humanoid robots to reproduce complex reference motions. DeepMimic demonstrated the tracking of highly dynamic motion clips~\cite{peng2018deepmimic}. AMP and PHC extended this capability toward more diverse and robust humanoid behaviors~\cite{peng2021amp,luo2023phc}. BeyondMimic further demonstrated reliable motion tracking on physical humanoid robots~\cite{beyondmimic}.

Recent methods also study transitions between robot motions. Switch connects similar states across motion clips through a skill graph and learns a policy that moves through the resulting paths~\cite{lau2026switch}. Perceptive Humanoid Parkour uses motion matching to select suitable entry states between locomotion and dynamic motions~\cite{wu2026perceptive}. RPG blends multiple expert policies to support transitions between fighting motions~\cite{rpg_2026}. These methods handle transitions through state selection
or policy learning. STAMP selects spatial and temporal boundary conditions for independently acquired motions and generates the connecting reference using a human motion prior.

\subsection{Human Motion Inbetweening}

Motion inbetweening generates missing motion from observed boundary contexts. Earlier learning based approaches use recurrent networks or Transformers to recover the missing interval~\cite{harvey2020robust,kim2022conditional,tang2022transition}. Later methods use phase representations or diffusion models to improve motion quality and control~\cite{starke2023phase,cohan2024flexible}. TEACH and PriorMDM generate temporally composed human motions from sequences of text descriptions~\cite{TEACH:3DV:2022,shafir2024human}.

Recent motion models support text, keyframe, trajectory, and online constraints~\cite{kimodo_2026,ardy_2026}. MotionBricks applies motion inbetweening to animation and real-world control of the Unitree~G1~\cite{motionbricks_2026}. These models generate motion under specified boundary conditions. When the boundaries are extracted from a continuous motion, their relative placement and timing are
already defined. STAMP addresses independently acquired motions by
selecting their relative position, heading, and transition duration before applying a motion inbetweening model.

\subsection{Human and Robot Motion Retargeting}

Motion retargeting transfers motion between bodies with different joint structures and proportions. Geometric retargeting uses joint correspondences and inverse kinematics to adapt human motion to a humanoid robot~\cite{darvish2019wholebody}. Learning based methods instead learn shared or cross embodiment motion representations~\cite{choi2020nonparametric,yan2023imitationnet,nmr_2026}. GMR provides a general motion retargeting pipeline for constructing robot motion from human motion data~\cite{gmr_2025}. Other methods use normalized limb directions and target body proportions to transfer motion across different kinematic structures~\cite{jeong2025rig,core_2025}.

STAMP adapts CoRe~\cite{core_2025} for bidirectional retargeting, enabling robot motions to be connected using a human motion prior and transferred back to
the target robot.

\section{Problem Formulation}
\label{sec:problem_formulation}

Let $\mathcal{M}_P^R$ and $\mathcal{M}_F^R$ denote two independently
acquired motions for robot $R$, containing $T_P$ and $T_F$ frames,
respectively. Each frame specifies the robot's root state and joint
configuration. Our goal is to connect the preceding motion
$\mathcal{M}_P^R$ to the following motion $\mathcal{M}_F^R$.

The motions are specified independently, so their relative position,
heading, and transition duration are not given. Each motion may be
plausible on its own, yet their boundary states may be difficult to
connect under an arbitrary placement. The problem therefore requires
determining both their spatial and temporal relationship and the
motion connecting them.

We seek a completed robot motion of the form
\begin{equation}
    \mathcal{M}_{P\rightarrow F}^R
    =
    \overline{\mathcal{M}}_P^R
    \oplus
    \mathcal{T}_{P\rightarrow F}^R
    \oplus
    \overline{\mathcal{M}}_F^R,
    \label{eq:robot_transition_sequence}
\end{equation}
where $\oplus$ denotes temporal concatenation,
$\mathcal{T}_{P\rightarrow F}^R$ is the transition, and
$\overline{\mathcal{M}}_P^R$ and $\overline{\mathcal{M}}_F^R$ are
the corresponding input-motion segments in the completed sequence. These segments may be rigidly transformed, but should retain the content and order of the input motions.

The transition should provide continuous body movement between the
two boundaries without requiring a return to a predefined neutral pose
or introducing an unintended action. The completed sequence serves
as a reference for robot motion tracking.

\section{Proposed Method}
\label{sec:method}

\begin{figure*}[!t]
    \centering
    \includegraphics[width=1.90\columnwidth]
    {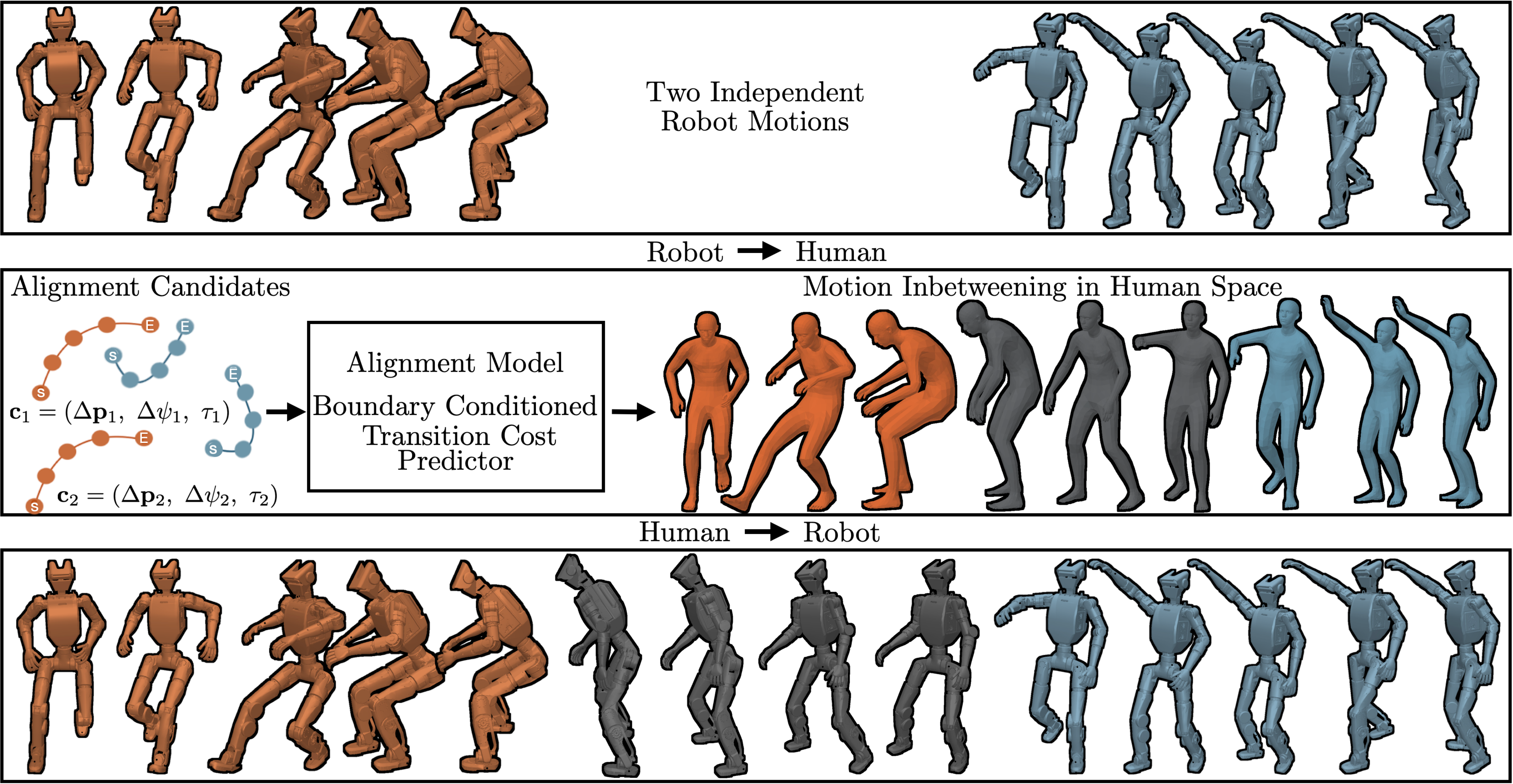}
    \caption{Overview of STAMP. Robot motions are transferred to a human rig. The alignment model predicts a transition cost for each candidate position, heading, and duration. The lowest-cost candidate is selected to condition motion inbetweening in human space. The completed sequence is retargeted to the robot and used as a motion-tracking reference.}
    \label{fig:overview}
    \vspace{-6pt}
\end{figure*}

STAMP performs alignment and transition generation in human motion space to reuse human motion knowledge across robot
embodiments. We use SOMA~\cite{soma2026} as the human rig and retarget the input robot motions to it, yielding
$\mathcal{M}_P^H$ and $\mathcal{M}_F^H$ with $T_P$ and $T_F$ frames, respectively. As shown in Fig.~\ref{fig:overview}, the alignment model predicts a transition cost for each candidate arrangement, which
specifies the following motion's position and heading and the transition duration. The candidate with the lowest predicted cost is selected, and a human motion prior generates the connecting motion under the selected conditions. The complete sequence is then retargeted to the robot and used to train a motion-tracking policy.

\subsection{Spatial and Temporal Motion Alignment}
\label{sec:motion_alignment}

An alignment candidate specifies the placement of the following
motion and the time available for the transition:
\begin{equation}
    \mathbf{c}
    =
    \left(
        \Delta\mathbf{p}_c,
        \Delta\psi_c,
        \tau_c
    \right),
    \label{eq:alignment_candidate}
\end{equation}
where $\Delta\mathbf{p}_c\in\mathbb{R}^{2}$ and $\Delta\psi_c$
specify the initial planar root position and heading of the
following motion relative to the final root frame of the
preceding motion. The transition duration $\tau_c$ is in seconds.

The candidate transforms the following motion as
\begin{equation}
    \widetilde{\mathcal{M}}_F^H(\mathbf{c})
    =
    \mathcal{A}
    \left(
        \mathcal{M}_F^H;
        \Delta\mathbf{p}_c,
        \Delta\psi_c
    \right),
    \label{eq:following_transform}
\end{equation}
where $\mathcal{A}$ places its initial root at the candidate
planar position and heading and applies the same rigid
transformation to the complete following motion.

\paragraph{Alignment model}

Let $K$ denote the number of frames in each boundary context.
The alignment inputs $\mathbf{Z}_P$ and $\mathbf{Z}_F(\mathbf{c})$
contain features from the final $K$ preceding frames and the
first $K$ transformed following frames, respectively.
These features describe root and body states, velocities,
and foot contacts. We additionally provide
$\boldsymbol{\rho}(\mathbf{c})$, which describes relative foot
placement and contact configuration across the boundaries.

The model predicts a scalar transition cost:
\begin{equation}
    e_{\phi}(\mathbf{c})
    =
    E_{\phi}
    \left(
        \mathbf{Z}_P,
        \mathbf{Z}_F(\mathbf{c}),
        \tau_c,
        \boldsymbol{\rho}(\mathbf{c})
    \right),
    \label{eq:predicted_transition_cost}
\end{equation}
where $E_{\phi}$ is the alignment network with parameters
$\phi$. A lower predicted cost indicates a more suitable
alignment candidate.

Separate temporal Transformer encoders process the two
contexts. Their features, difference, and element-wise product
are combined with the duration and relative foot configuration.
An MLP maps this representation to the transition cost.
Temporal contexts allow the model to consider incoming and
outgoing motion dynamics beyond the endpoint poses.

\paragraph{Alignment supervision}

We divide each continuous training motion into a preceding
context, an observed transition $\mathcal{M}_B^H$, and a
following context. For candidate $\mathbf{c}$, the observed
transition is resampled to duration $\tau_c$, and the following
context is rigidly transformed. We construct
\begin{equation}
\begin{aligned}
    \mathbf{X}(\mathbf{c})
    ={}&
    \left[\mathbf{x}_{P,\mathrm{end}}^H\right]
    \oplus
    \mathcal{R}_{\tau_c}(\mathcal{M}_B^H)\\
    &\oplus
    \left[
        \widetilde{\mathbf{x}}_{F,\mathrm{start}}^H(\mathbf{c})
    \right],
\end{aligned}
\label{eq:counterfactual_transition}
\end{equation}
where $\mathbf{x}_{P,\mathrm{end}}^H$ and
$\widetilde{\mathbf{x}}_{F,\mathrm{start}}^H(\mathbf{c})$
are the last preceding frame and first transformed following
frame. Brackets indicate single-frame sequences, $\oplus$
denotes temporal concatenation, and $\mathcal{R}_{\tau_c}$
denotes temporal resampling.

Candidate durations correspond to 15, 30, 45, or 60 frames
at 30 fps. The transition channels are linearly interpolated,
with heading angles unwrapped before interpolation.
Contact channels are also interpolated and treated as soft
weights. The transition is retimed but is not spatially
warped toward the following context. We compute finite
differences over the complete sequence $\mathbf{X}(\mathbf{c})$
at 30 fps, so boundary mismatches contribute to the
kinematic costs. The observed transition is used only
for supervision and is never provided to the alignment model.

For a quantity $\mathbf{x}$ and normalization scale $\sigma$,
we define
\begin{equation}
    \mathcal{Q}(\mathbf{x};\sigma)
    =
    \operatorname{mean}_{t,d}
    \left[
        \left(\frac{x_{t,d}}{\sigma}\right)^2
    \right],
    \label{eq:normalized_mean_square}
\end{equation}
where the mean includes all valid timesteps and scalar
feature dimensions. Table~\ref{tab:supervision_cost}
defines the cost components and their combination weights.
Candidate dependence is omitted within the table for brevity.

\begin{table}[t]
\centering
\small
\setlength{\tabcolsep}{3pt}
\caption{Kinematic supervision costs and combination weights.
Normalization scales use the corresponding SI units.}
\label{tab:supervision_cost}
\begin{tabular}{@{}llc@{}}
\toprule
Term & Definition & Weight \\
\midrule
$E_{\mathrm{root}}$
& $\mathcal{Q}(\ddot{\mathbf{p}}_r;3)
 +0.25\mathcal{Q}(\dddot{\mathbf{p}}_r;25)$
& 1.0 \\

$E_{\mathrm{body}}$
& $\mathcal{Q}(\ddot{\mathbf{p}}_b;6)
 +0.25\mathcal{Q}(\dddot{\mathbf{p}}_b;50)$
& 0.6 \\

$E_{\mathrm{foot}}$
& $\mathcal{Q}(\ddot{\mathbf{p}}_f;5)
 +0.25\mathcal{Q}(\dddot{\mathbf{p}}_f;45)$
& 0.8 \\

$E_{\mathrm{stance}}$
& $\mathcal{Q}(\dot{\mathbf{s}};1)
 +0.25\mathcal{Q}(\ddot{\mathbf{s}};6)$
& 0.6 \\

$E_{\mathrm{yaw}}$
& $\mathcal{Q}(\dot{\psi};\pi)
 +0.5\mathcal{Q}(\ddot{\psi};2\pi)$
& 0.5 \\
& ${}+0.1\mathcal{Q}(\dddot{\psi};8\pi)$ & \\

$E_{\mathrm{skate}}$
& $\mathcal{Q}(\widetilde{\mathbf{c}}
 \odot\mathbf{v}^{xy};0.35)$
& 0.8 \\

$E_{\mathrm{idle}}$
& $g_{\mathrm{idle}}(\tau_c/0.5)^2$
& 1.0 \\
\bottomrule
\end{tabular}
\end{table}

Here, $\mathbf{p}_r$ and $\mathbf{p}_b$ denote the root position and all SOMA joint positions, including the root.
We reconstruct global joint positions by adding the root translation to the stored root-translation-relative positions.
Before temporal differencing, all positions are expressed in a common canonical frame defined by the preceding
motion's final root position and heading. Candidate displacements and heading differences are preserved in this frame. The foot-dynamics term uses two foot
centers $\mathbf{p}_f$, each defined as the mean of its foot and toe-base joint positions. The stance vector $\mathbf{s}$ points from the left foot center to the right, and $\psi$ is the unwrapped root heading. The skating term uses the horizontal velocities $\mathbf{v}^{xy}$ of the four foot and toe-base sites. Their soft contact weights $\widetilde{\mathbf{c}}$ are applied to each velocity coordinate using element-wise
multiplication $\odot$.

To penalize long transitions with little motion, we define:
\begin{equation}
\begin{aligned}
    a &= v_{\mathrm{body}}
       +0.5v_{\mathrm{root}}+0.15v_{\mathrm{yaw}},\\
    g_{\mathrm{idle}}
      &= \operatorname{clip}\left(1-a/0.12,0,1\right),
\end{aligned}
\label{eq:idle_gate}
\end{equation}
where the activity terms are the RMS body linear, root
linear, and heading velocities, respectively. Each RMS
value is computed over all scalar velocity components
obtained by finite differences over the complete
boundary-inclusive sequence $\mathbf{X}(\mathbf{c})$,
including both boundary-crossing intervals.

Let $E_k(\mathbf{c})$ denote component $k$ and $\lambda_k$
its weight. The target transition cost is
\begin{equation}
    e^{*}(\mathbf{c})
    =
    \log
    \left(
        1+\sum_k \lambda_k E_k(\mathbf{c})
    \right).
    \label{eq:target_transition_cost}
\end{equation}
This target provides a kinematic proxy for alignment selection rather than a direct estimate of the cost of a generated transition.

The logarithm compresses large cost values. The normalization scales and weights were chosen heuristically based on the units and expected magnitudes of the kinematic quantities. They were not tuned using motion-inbetweening outputs, policy rewards, or robot-tracking results.

For each training example, $\mathbf{e}^{*}$ and
$\mathbf{e}_{\phi}$ collect the target and predicted costs in the same candidate order. Their preference distributions are
\begin{equation}
\begin{aligned}
    \mathbf{q}^{*}
    &= \operatorname{softmax}(-\mathbf{e}^{*}/T_t),\\
    \mathbf{q}_{\phi}
    &= \operatorname{softmax}(-\mathbf{e}_{\phi}/T_p),
\end{aligned}
\label{eq:alignment_distributions}
\end{equation}
where $T_t$ and $T_p$ are the target and prediction temperatures. We train the model using
\begin{equation}
\begin{aligned}
    \mathcal{L}_{\mathrm{align}}
    ={}&
    D_{\mathrm{KL}}
    \left(
        \mathbf{q}^{*}
        \,\middle\|\,
        \mathbf{q}_{\phi}
    \right)\\
    &+
    \lambda_{\mathrm{reg}}
    \operatorname{SmoothL1}
    \left(
        \mathbf{e}_{\phi},
        \mathbf{e}^{*}
    \right),
\end{aligned}
\label{eq:alignment_loss}
\end{equation}
where $\lambda_{\mathrm{reg}}$ weights the regression term. The KL-divergence term learns candidate preferences, while
regression anchors the predicted costs to their targets. We use $T_t=0.5$, $T_p=0.15$, and $\lambda_{\mathrm{reg}}=0.35$.

\subsection{Particle-Based Inference}
\label{sec:particle_inference}

STAMP constructs a fixed set $\mathcal{C}$ of alignment particles by combining predefined planar positions, headings, and transition durations relative to the final root frame of the preceding motion.
The candidate values are provided in
Sec.~\ref{sec:experimental_setup}.

The alignment model evaluates all particles without generating their
corresponding transitions. STAMP selects
\begin{equation}
    \mathbf{c}^{*}
    =
    \underset{\mathbf{c}\in\mathcal{C}}{\arg\min}
    \;e_{\phi}(\mathbf{c}).
    \label{eq:particle_selection}
\end{equation}
The selected placement, heading, and duration are then supplied to the human motion inbetweening model.

\subsection{Human Motion Inbetweening}
\label{sec:human_prior}

We develop a human motion inbetweening model to generate the transition under the selected alignment. In this subsection,
$\widetilde{\mathcal{M}}_F^H$ denotes the following motion placed according to $\mathbf{c}^{*}$, and $\tau$ denotes its selected transition duration.

The model observes the final $K$ frames of the preceding motion and the first $K$ frames of the aligned following motion:
\begin{equation}
\begin{aligned}
    \mathbf{X}_P
    &=
    \left[
        \mathbf{x}_{P,T_P-K+1}^{H},
        \ldots,
        \mathbf{x}_{P,T_P}^{H}
    \right],\\
    \mathbf{X}_F
    &=
    \left[
        \widetilde{\mathbf{x}}_{F,1}^{H},
        \ldots,
        \widetilde{\mathbf{x}}_{F,K}^{H}
    \right].
\end{aligned}
\label{eq:inbetweening_contexts}
\end{equation}
Here, $\mathbf{x}_{P,t}^{H}$ and
$\widetilde{\mathbf{x}}_{F,t}^{H}$ are frames of
$\mathcal{M}_P^H$ and $\widetilde{\mathcal{M}}_F^H$, respectively. The transition contains
$N_\tau=\operatorname{round}(f_s\tau)$ frames, where $f_s=30$ fps.

Each frame is represented as
\begin{equation}
    \mathbf{x}_t^H
    =
    \left[
        \mathbf{r}_t,
        \mathbf{p}_t,
        \mathbf{R}_t,
        \mathbf{v}_t,
        \mathbf{f}_t
    \right],
    \label{eq:human_motion_representation}
\end{equation}
where $\mathbf{r}_t$ represents root motion, $\mathbf{p}_t$ and $\mathbf{R}_t$ represent joint positions and rotations, and $\mathbf{v}_t$ and $\mathbf{f}_t$ represent joint velocities and foot contacts.

Our diffusion model adopts the motion representation and two-stage Transformer design of Kimodo~\cite{kimodo_2026}. It is trained for motion inbetweening using boundary contexts without text conditioning. During denoising, the boundary frames remain observed, and the
intermediate frames are generated.

At each denoising step, the first Transformer predicts the three-dimensional root position and the sine and cosine of its heading. These predictions are converted to a local representation containing heading velocity, planar velocity, and root height. The second Transformer predicts articulated body motion conditioned on this root representation and the boundary contexts.
The generated intermediate frames form the human transition
$\mathcal{T}_{P\rightarrow F}^H$.

We train on continuous human motion windows of length $2K+N_{\tau}$. The first and last $K$ frames are observed conditions, and the intermediate frames are the generation target. Training samples use different durations and are padded to a common batch length. Padded
frames are excluded from attention and loss computation. We apply reconstruction losses only to the unobserved transition frames, with channel weights specified in
Sec.~\ref{sec:experimental_setup}.

The models are trained separately, allowing the selected alignment to condition another compatible inbetweening model without retraining the alignment network.

\subsection{Robot-to-Human-to-Robot Motion Retargeting}
\label{sec:motion_retargeting}

Retargeting connects the robot-specific motion representation with the human motion space used for alignment and generation. We denote the two directions by $\mathcal{P}_{R\rightarrow H}$ and
$\mathcal{P}_{H\rightarrow R}$ and implement both by adapting CoRe~\cite{core_2025}.

The input robot motions are transferred to the human rig as
\begin{equation}
\begin{aligned}
    \mathcal{M}_P^H
    &=
    \mathcal{P}_{R\rightarrow H}(\mathcal{M}_P^R),\\
    \mathcal{M}_F^H
    &=
    \mathcal{P}_{R\rightarrow H}(\mathcal{M}_F^R).
\end{aligned}
\label{eq:robot_to_human}
\end{equation}
To obtain motion compatible with the human prior, we adapt root displacement and foot placement to human body proportions while retaining the motion content, heading, and support-foot sequence. Human joints without robot correspondences are maintained in temporally consistent configurations.

After transition generation, the complete sequence is transferred back to the target robot:
\begin{equation}
\begin{aligned}
    \mathcal{M}_{P\rightarrow F}^{R}
    =
    \mathcal{P}_{H\rightarrow R}
    \Big(
        \mathcal{M}_P^H
        \oplus
        \mathcal{T}_{P\rightarrow F}^H
        \oplus
        \widetilde{\mathcal{M}}_F^H
    \Big).
\end{aligned}
\label{eq:human_to_robot}
\end{equation}
Human-to-robot retargeting accounts for the target body proportions, joint structure, joint limits, and foot contacts. We retarget the connected motion as a single sequence to maintain consistency across the transition boundaries. The resulting robot motion provides the reference for subsequent motion-tracking policy training and execution.

\section{Experiments}
\label{sec:experiments}

We compare STAMP with interpolation and inbetweening baselines on Unitree~G1 and ROBOTIS~K1 under the same tracking-policy training protocol. We also evaluate alignment selection and demonstrate long-horizon motion composition on the physical robot, ROBOTIS~K1.

\subsection{Experimental Setup}
\label{sec:experimental_setup}

\paragraph{Motion pairs}

Our benchmark contains 18 motion pairs across three scenarios: eight from boxing, five from industrial maintenance, and five from security and traffic control. The pairs include changes in locomotion, heading, stance height, foot placement, and upper-body movement.

Each constituent motion is independently generated using Kimodo~\cite{kimodo_2026} at 30 fps. Each prompt describes a single
motion without specifying its preceding or following motion. The generated motions are approximately 6--9 seconds long. We remove initialization and terminal settling when necessary. The generated human motions are retargeted to Humanoids using the human-to-robot retargeting procedure described in Sec.~\ref{sec:motion_retargeting}. All methods use the same input motion pairs for each robot.

\paragraph{Compared methods}

\textbf{Root/heading interpolation} matches the following motion's initial root position and heading to the preceding motion's final root frame and connects the boundary states using quintic interpolation.

\textbf{Kimodo$^{\dagger}$} uses Kimodo's motion representation and denoising architecture~\cite{kimodo_2026} and is trained for inbetweening with boundary motion contexts instead of text. It uses the same generator and Robot-to-Human-to-Robot pipeline as STAMP, but retains the original boundary arrangement.

\textbf{MotionBricks} uses its released Unitree~G1 model and duration predictor~\cite{motionbricks_2026} to generate transitions under the original boundary arrangement. We evaluate MotionBricks only on Unitree~G1, for which an official pretrained model is available.

\textbf{STAMP} selects the following motion's position, heading, and transition duration before generation with Kimodo$^{\dagger}$. Interpolation and Kimodo$^{\dagger}$ use a fixed duration of $1.0$\,s. The comparison evaluates spatial alignment and duration selection together.

\paragraph{Alignment particles}

Relative to the preceding motion's final root frame, STAMP combines planar offsets $\{-2,-2/3,2/3,2\}$\,m along each ground-plane axis, headings $\{-60,-30,0,30,60,90\}^{\circ}$, and durations $\{0.5,1.0,1.5,2.0\}$\,s. The Cartesian product yields 384 candidates. Adding the identity placement $(0,0,0^{\circ})$ for each duration gives 388 particles per pair. STAMP selects the particle with the lowest predicted cost, while the ablation uses the rank-100 particle from the same set.

\paragraph{Model training}

Both the alignment model and Kimodo$^{\dagger}$ are trained on BONES-SEED~\cite{bonesseed} using the SOMA representation at 30 fps with $K=10$ boundary frames. Each alignment sample contains 16 candidates spanning observed and identity placements, alternative durations, and local and global spatial perturbations. The alignment model uses two three-layer Transformer encoders of width 160 with five attention heads. It is trained for 25,000 steps using AdamW with learning rate $2\times10^{-4}$ and per-GPU batch size 32. We select the checkpoint with the lowest validation loss.

Kimodo$^{\dagger}$ uses Transformer width 512 with eight attention heads, 1,000 diffusion timesteps, and 50 denoising steps at inference. Smooth L1 loss weights for root position, heading, joint position, rotation, velocity, and contact are $(10,2,10,10,3,4)$, respectively.
Training uses AdamAtan2 with peak learning rate $2\times10^{-5}$, 2,000 warm-up steps, cosine decay, and global batch size 2,048.

\paragraph{Motion tracking}

Each reference is tracked by an independently trained BeyondMimic policy~\cite{beyondmimic}. Pairwise references contain the transition and $K$ context frames on each side, resampled from 30 to 50 Hz. Training starts from randomly sampled reference phases.

Each policy is trained for 5,000 iterations with 2,048 parallel environments: 1,500 iterations without external pushes, 2,000 with moderate perturbations and domain randomization, and 1,500 with full deployment randomization. Architecture, observations, rewards, termination conditions, and randomization schedules are identical across methods for each robot. We use one training seed per reference and evaluate the final checkpoint.

\subsection{Evaluation Metrics}
\label{sec:evaluation_metrics}

Each policy is evaluated over 1,024 rollouts using 256 parallel environments. We discard the initial rollout of each environment and collect subsequent rollouts after regular resets. Each evaluation rollout starts at the first
reference frame, with randomized perturbations to the corresponding root pose, root velocity, and joint configuration. The reference phase is not randomized.

Body velocity error and joint acceleration are averaged over valid timesteps up to completion or early termination,
including both successful and failed episodes. We then average equally over 1,024 episodes per motion pair and
over the 18 pairs, totaling 18,432 evaluation rollouts per method.

\paragraph{Success rate}

A rollout is successful if it reaches the final reference frame without triggering a tracking termination and maintains
root-position, root-orientation, and end-effector-position errors below $0.20$\,m, $0.40$\,rad, and $0.20$\,m,
respectively, throughout execution. The end-effector condition applies to every tracked end effector. Success rate is the
percentage of successful rollouts.

\paragraph{Body velocity error}

Mean body velocity error ($E_{\mathrm{vel}}$) is the mean
Euclidean difference between simulated and reference world-frame linear velocities over all 30 tracked body points and evaluated timesteps. The root body is included, and root velocity is not
subtracted. We report $E_{\mathrm{vel}}$ in m/s.

\paragraph{Joint acceleration}

We report the mean absolute acceleration of the executed actuated joints:
\begin{equation}
E_{\mathrm{acc}}
=
\frac{1}{(T-1)D}
\sum_{t=2}^{T}
\sum_{d=1}^{D}
\left|
\frac{\dot{q}_{t,d}-\dot{q}_{t-1,d}}{\Delta t}
\right|,
\label{eq:mean_joint_acceleration}
\end{equation}
where $\dot{q}_{t,d}$ is the velocity of actuated joint $d$
at timestep $t$, $T$ is the number of evaluated timesteps,
$D$ is the number of actuated joints, and $\Delta t=0.02$\,s.
The floating root is excluded. Lower values indicate smaller changes in executed joint velocity. We interpret this metric together with success rate and velocity error, since low acceleration alone does not establish accurate tracking.

Velocity error and joint acceleration are evaluated over the transition clip, including its short preceding and following
boundary contexts.

\subsection{Pairwise Motion Transition}
\label{sec:pairwise_evaluation}

Tables~\ref{tab:g1_tracking} and~\ref{tab:k1_tracking} report
tracking performance over 18 motion pairs. STAMP uses the same human motion inbetweening model and alignment model for both robots, with robot-specific retargeting and tracking policies.

\begin{table}[t]
\centering
\small
\setlength{\tabcolsep}{3.2pt}
\renewcommand{\arraystretch}{1.08}
\caption{Unitree~G1 transition tracking over 18 motion pairs. Kimodo$^{\dagger}$ denotes the Kimodo-based inbetweening model.}
\label{tab:g1_tracking}
\begin{tabular}{@{}lccc@{}}
\toprule
Method
& \shortstack{Success Rate\\(\%) $\uparrow$}
& \shortstack{$E_{\mathrm{vel}}$\\(m/s) $\downarrow$}
& \shortstack{$E_{\mathrm{acc}}$\\(rad/s$^2$) $\downarrow$} \\
\midrule
Root/heading interpolation
& 69.59 & 0.2581 & 13.57 \\
MotionBricks
& 56.91 & 0.3433 & 17.67 \\
Kimodo$^{\dagger}$
& 50.69 & 0.5293 & 18.98 \\
\midrule
\textbf{STAMP}
& \textbf{80.69} & \textbf{0.2289} & \textbf{12.43} \\
\bottomrule
\end{tabular}%
\end{table}

\begin{table}[t]
\centering
\small
\setlength{\tabcolsep}{3.2pt}
\renewcommand{\arraystretch}{1.08}
\caption{ROBOTIS~K1 transition tracking over 18 motion pairs.
Kimodo$^{\dagger}$ denotes the Kimodo-based inbetweening model.}
\label{tab:k1_tracking}
\begin{tabular}{@{}lccc@{}}
\toprule
Method
& \shortstack{Success Rate\\(\%) $\uparrow$}
& \shortstack{$E_{\mathrm{vel}}$\\(m/s) $\downarrow$}
& \shortstack{$E_{\mathrm{acc}}$\\(rad/s$^2$) $\downarrow$} \\
\midrule
Root/heading interpolation
& 84.70 & 0.1964 & 11.74 \\
Kimodo$^{\dagger}$
& 64.55 & 0.5303 & 16.50 \\
\midrule
\textbf{STAMP}
& \textbf{87.52} & \textbf{0.1860} & \textbf{11.12} \\
\bottomrule
\end{tabular}%
\end{table}

STAMP achieves the highest success rate on both robots, reaching $80.69\%$ on Unitree~G1 and $87.52\%$ on ROBOTIS~K1. It also yields the lowest body velocity error and mean executed joint acceleration among the evaluated methods.

The comparison with Kimodo$^{\dagger}$ supports the benefit of selecting
spatial and temporal boundary conditions before generation using the same motion prior. MotionBricks uses a different
generator, representation, and duration model, so its comparison with STAMP evaluates complete transition systems.

Each policy is evaluated against its own reference. These results therefore measure the trackability of the generated
transitions under a common training protocol, rather than similarity to a shared ground-truth transition.

\subsection{Alignment Ranking Ablation}
\label{sec:alignment_ranking_ablation}

We compare candidates ranked 1 and 100 in ascending predicted transition cost. Both conditions use the same candidate set,
motion generator, retargeting procedure, and tracking-policy training and evaluation protocol. Only the selected position, heading, and transition duration differ.

\begin{table}[t]
\centering
\small
\setlength{\tabcolsep}{3.2pt}
\renewcommand{\arraystretch}{1.08}
\caption{Tracking performance at alignment ranks 1 and 100.
Rank 1 has the lowest predicted transition cost.}
\label{tab:alignment_ranking_ablation}
\begin{tabular}{@{}llccc@{}}
\toprule
Robot & Candidate
& \shortstack{Success Rate\\(\%) $\uparrow$}
& \shortstack{$E_{\mathrm{vel}}$\\(m/s) $\downarrow$}
& \shortstack{$E_{\mathrm{acc}}$\\(rad/s$^2$) $\downarrow$} \\
\midrule
Unitree~G1 & Rank 1
& \textbf{80.69} & \textbf{0.2289} & \textbf{12.43} \\
& Rank 100
& 71.33 & 0.3001 & 15.97 \\
\midrule
ROBOTIS~K1 & Rank 1
& \textbf{87.52} & \textbf{0.1860} & \textbf{11.12} \\
& Rank 100
& 71.90 & 0.2875 & 14.34 \\
\bottomrule
\end{tabular}%

\end{table}

As shown in Table~\ref{tab:alignment_ranking_ablation},
rank 1 achieves higher success rates than rank 100 by 15.62 percentage points on ROBOTIS~K1 and 9.36 percentage points on Unitree~G1. It produces lower body velocity error and executed joint acceleration on both robots. These results support the usefulness of the learned transition cost for selecting trackable boundary arrangements in this comparison.

\subsection{Physical Robot Evaluation}
\label{sec:physical_evaluation}

We execute complete motion sequences on the physical version of ROBOTIS~K1 using the same motion-tracking framework. These sequences include the constituent motions and their connecting transitions.

\paragraph{Running-to-punching transition}

\begin{figure}[t]
\centering
\includegraphics[width=\columnwidth]
{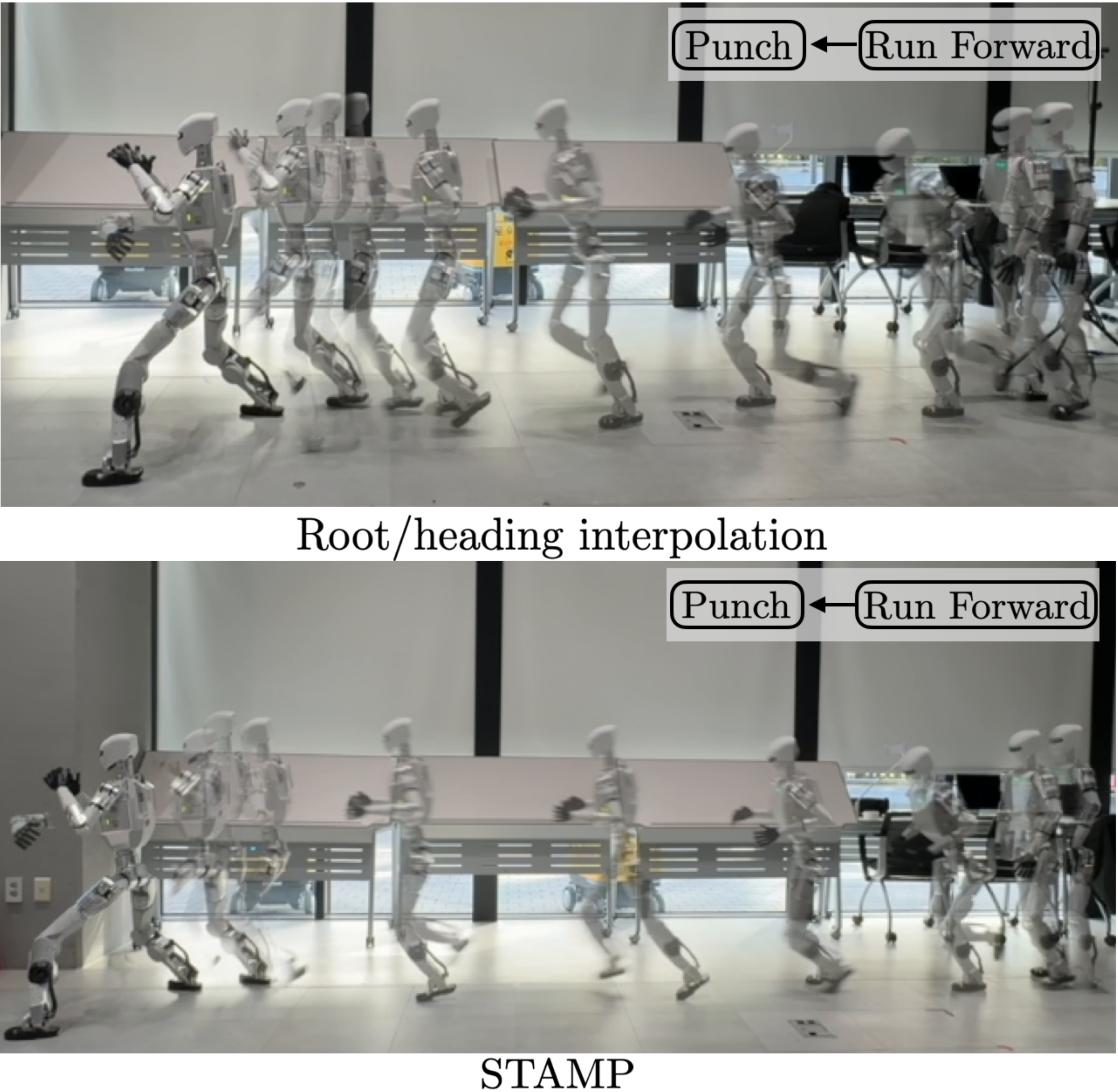}
\caption{Running-to-punching transitions on ROBOTIS~K1 using interpolation (top) and STAMP (bottom), progressing from right to left. STAMP places the punching motion farther forward, providing space for a connecting step during the transition.}
\label{fig:physical_comparison}
\vspace{-6pt}
\end{figure}

Fig.~\ref{fig:physical_comparison} compares root/heading interpolation and STAMP using the same running and punching
motions. Interpolation places the punching start at the running endpoint, leaving no net forward displacement between the transition boundaries. During physical execution, we observe foot sliding as the robot slows down under interpolation. STAMP places the punching motion farther
along the running direction, and the inbetweening model generates a connecting step that allows continued forward
movement during the transition.

\paragraph{Motor-current response}

We examine the motor-current response when connecting a look-around motion and a stop gesture in a traffic-control scenario. Both methods use a 1.5\,s transition and are executed five times on ROBOTIS~K1. We extract the motor-current signals reported by the robot driver at 200\,Hz, apply a fourth-order zero-phase low-pass filter with a 20\,Hz cutoff, and compute the RMS across 23 motors at each timestep. Values are retained in driver-reported units without conversion to amperes.

\begin{figure}[t]
\centering
\includegraphics[width=\columnwidth]
{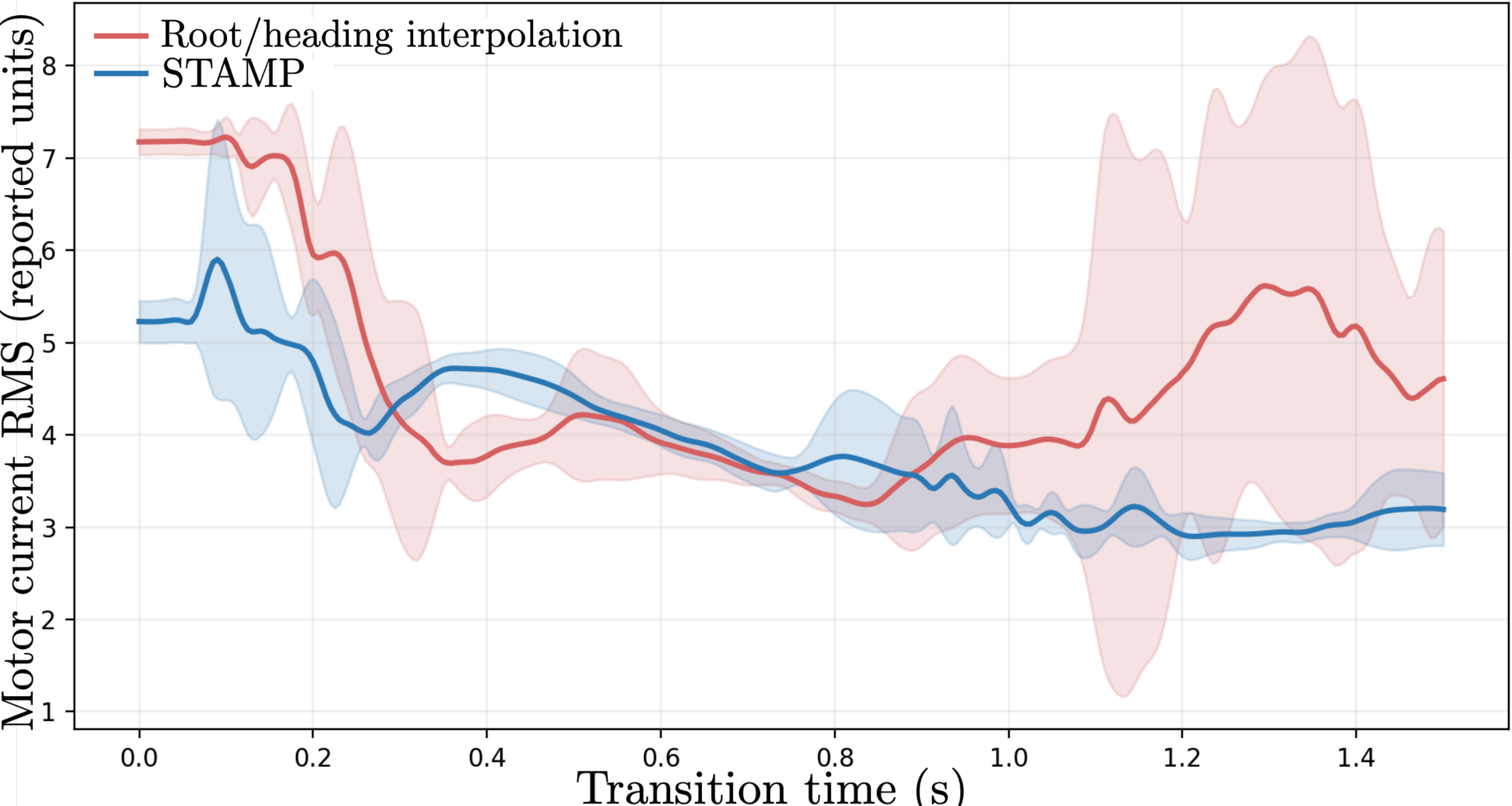}
\caption{Motor-current response during a transition from looking around to making a stop gesture on ROBOTIS~K1. Both methods use a 1.5\,s transition. Lines and shading indicate the mean and $\pm1$ standard deviation across five executions of the RMS signal over 23 motors. Time zero denotes transition onset. Values are in driver-reported units.}
\label{fig:current_response}
\end{figure}

Fig.~\ref{fig:current_response} shows the mean and standard deviation across executions aligned at transition onset.
For this motion pair, STAMP exhibits a lower mean RMS signal and generally smaller between execution variation
in the latter part of the transition. The current levels already differ at transition onset, so this comparison
describes the observed execution responses rather than isolating the effect of alignment alone.

\paragraph{Long-horizon composition}

We repeatedly apply STAMP to connect independently acquired motions into a longer sequence. At each connection, STAMP
selects the alignment candidate with the lowest predicted transition cost and generates the connecting motion.

\begin{figure*}[t]
\centering
\includegraphics[width=1.90\columnwidth]
{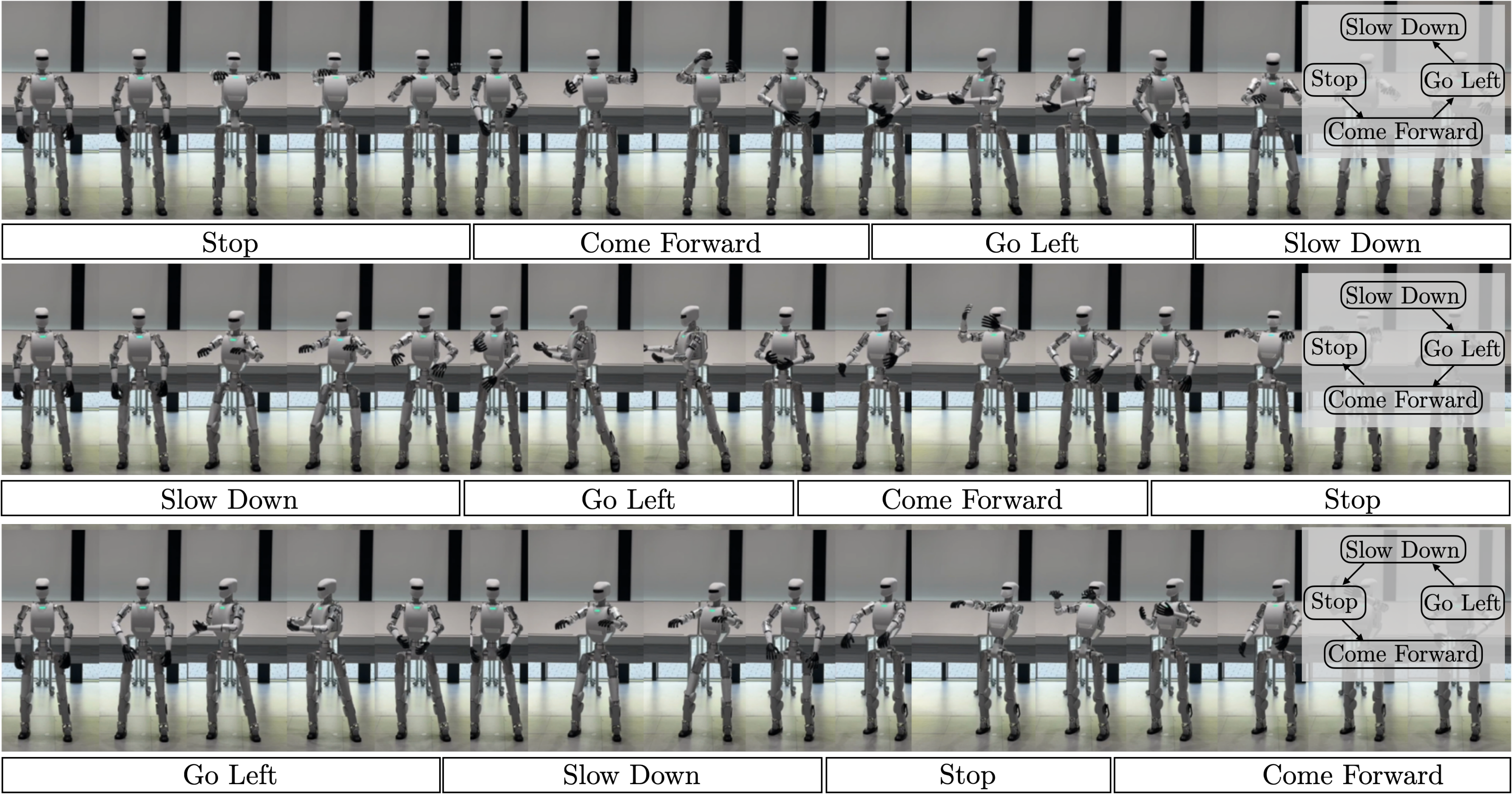}
\caption{Long-horizon compositions of the same four
traffic-control motions executed on the ROBOTIS~K1. Each row shows a different composition. STAMP selects the spatial and temporal alignment and generates the transitions between consecutive motions.}
\label{fig:motion_composition}
\vspace{-6pt}
\end{figure*}

Fig.~\ref{fig:motion_composition} shows three compositions of the same four traffic-control motions. These real robot executions demonstrate how pairwise alignment and transition generation can compose independently acquired motions into longer sequences.

\subsection{Limitations}
\label{sec:limitations}

The learned cost provides a kinematic guide for alignment selection, while downstream trackability also depends on retargeting quality and robot dynamics.

We currently train a tracking policy for each composed sequence. Future work could integrate alignment and inbetweening as motion priors within a shared reinforcement learning framework, allowing new compositions to be executed with less sequence-specific training.

\section{Conclusion}
\label{sec:conclusion}

We propose STAMP, a Robot-to-Human-to-Robot framework for connecting independently acquired humanoid motions.
A learned alignment model selects the relative position, heading, and transition duration before a human motion prior generates the transition. Both models operate in human motion space and are reused across robot embodiments. The completed sequence is retargeted and tracked by a robot-specific policy.

Experiments on Unitree~G1 and ROBOTIS~K1 show higher tracking success and lower body velocity errors than the evaluated baselines. The ablation supports learned
alignment selection, and physical demonstrations show connecting steps and longer motion compositions. A traffic-control case study also shows lower reported motor-current RMS than interpolation in the latter part of the transition.

\bibliographystyle{IEEEtran}
\bibliography{references}

\end{document}